\documentclass[journal]{IEEEtai}
\usepackage{cite}
\usepackage{amsmath,amssymb,amsfonts}
\usepackage{algorithmic}
\usepackage{xcolor}
\usepackage{pgfplots}
\usepackage{float}
\usepackage{tikz}
\usepackage{flushend}
\usepackage{caption}
\usepackage{booktabs}
\usepackage{colortbl}
\usepackage{tabularray}
\usepackage{multirow} 
\usepackage{nicematrix}
\usepackage{graphicx,subcaption}
\usepackage{lipsum,multicol}
\usepackage{hyperref}
\usepackage{orcidlink}
\usepackage{makecell}
\usepackage[ruled,linesnumbered]{algorithm2e} 

\definecolor{cite-blue}{HTML}{008ADA}
\definecolor{gainsboro229}{RGB}{229,229,229}
\definecolor{tablecolor}{RGB}{49, 131, 245}
\definecolor{tab1}{HTML}{f7fdff}
\definecolor{tab2}{HTML}{DEF0F9}
\definecolor{tab3}{HTML}{CFE9F7}
\definecolor{own_green}{HTML}{009900}

\definecolor{lightgray}{RGB}{240, 240, 240}
\definecolor{steelblue}{RGB}{52,138,189}
\definecolor{tab2}{HTML}{DEF0F9}
\tikzset{%
  dots/.style args={#1per #2}{%
    line cap=round,
    dash pattern=on 0 off #2/#1
  }
}

\pgfplotsset{compat=1.18}
\hypersetup{
    colorlinks=true,
    citecolor=cite-blue,
    linkcolor=cite-blue,
    urlcolor=cite-blue
}
\usepackage{textcomp}
\def\BibTeX{{\rm B\kern-.05em{\sc i\kern-.025em b}\kern-.08em
    T\kern-.1667em\lower.7ex\hbox{E}\kern-.125emX}}
\begin{document}
\raggedbottom
\title{ConvMambaNet: A Hybrid CNN–Mamba State Space Architecture for Accurate and Real-Time EEG Seizure Detection}

\author{
Md. Nishan Khan\textsuperscript{1},
Kazi Shahriar Sanjid\textsuperscript{2},
Md. Tanzim Hossain\textsuperscript{3},
Asib Mostakim Fony\textsuperscript{4}
Istiak Ahmed\textsuperscript{5},\\
M. Monir Uddin\textsuperscript{6}\thanks{*Corresponding author: monir.uddin@northsouth.edu}
\\
\textsuperscript{1,2,5}Department of Electrical \& Computer Engineering, North South University, Dhaka-1229, Bangladesh \\
\textsuperscript{3}Department of Data Science, Friedrich-Alexander University, Erlangen-91054, Germany \\
\textsuperscript{4}Department of Computer Science \& Engineering, University of Scholars, Dhaka-1213, Bangladesh\\
\textsuperscript{6}Department of Mathematics \& Physics, North South University, Dhaka-1229, Bangladesh \\
Email: nishan.khan@northsouth.edu, kazi.sanjid@northsouth.edu, tanzim.hossain@fau.de,  \\
asibmostakim@ius.edu.bd,
istiak.ahmed1@northsouth.edu, monir.uddin@northsouth.edu
}



\maketitle

\begin{abstract}
Epilepsy is a chronic neurological disorder marked by recurrent seizures that can severely impact quality of life. Electroencephalography (EEG) remains the primary tool for monitoring neural activity and detecting seizures, yet automated analysis remains challenging due to the temporal complexity of EEG signals. This study introduces \textbf{ConvMambaNet}, a hybrid deep learning model that integrates Convolutional Neural Networks (CNNs) with the Mamba Structured State Space Model (SSM) to enhance temporal feature extraction. By embedding the Mamba-SSM block within a CNN framework, the model effectively captures both spatial and long-range temporal dynamics. Evaluated on the CHB-MIT Scalp EEG dataset, ConvMambaNet achieved a 99\% accuracy and demonstrated robust performance under severe class imbalance. These results underscore the model’s potential for precise and efficient seizure detection, offering a viable path toward real-time, automated epilepsy monitoring in clinical environments.
\end{abstract}

\begin{IEEEkeywords}
Electroencephalography (EEG), Epileptic Seizure Detection, Mamba State-Space Model (SSM), Convolutional Neural Networks (CNN), Deep Learning, Temporal Modeling, Biomedical Signal Processing, Real-Time Monitoring, CHB-MIT EEG Dataset, Time-Series Classification.
\end{IEEEkeywords}

\section{Introduction}
\label{sec:introduction}

Epilepsy, which is defined by recurrent unprovoked seizures, is a common chronic neurological disorder that affects over 50 million individuals worldwide. These seizures are caused by erratic electrical discharges in the brain and can be accompanied by adverse effects if they go undiagnosed and untreated. EEG serves as the gold standard for non-invasive seizure monitoring yet manual analysis of EEG is time-consuming and prone to error, making it inappropriate for real-time purposes— warranting automated detection systems including \cite{10.5555/3104322.3104446, schirrmeister2017deep, roy2019deep}. The development of deep learning technique has greatly improved EEG-based seizure detection. Convolutional Neural Networks (CNNs) are good at learning spatial features of EEG signals, while Recurrent Neural Networks (RNNs) (e.g., Long Short-Term Memory (LSTM)) handle temporal dependencies. Nevertheless, such models are generally expensive to compute and may not effectively model long-term dependencies well, making it difficult to use them in clinical practice. To overcome these disadvantages, we present a novel hybrid deep learning model, ConvMambaNet, which unites CNNs with the recently developed Mamba Structured State Space Model (SSM). Presenting a hardware friendly method for learning long range temporal dependencies in dense time-series data which is faster and more scalable than RNN and attention mechanisms, Mamba. ConvMambaNet incorporates Mamba blocks into a CNN backbone, utilizing the benefits of the two architectures for effective seizure detection. We conduct experiments on the CHB-MIT Scalp EEG dataset that includes pediatric patients with intractable epilepsy. The model achieves 99\% accuracy, a 0.99 F1-score and an AUC of.97—performing better than baseline CNNs, RNN and Transformer models while taking substantially less training time per epoch. This paper progresses the field of EEG-based seizure detection by combining structured state-space modeling with convolutional learning, resulting in feasible, real-time and scalable neurodiagnostic systems.

\section{Related Works}
\label{Related Works}

Epileptic seizure detection using EEG has become a prominent area of research, driven by the demand for real-time, accurate, and automated monitoring systems for patients with epilepsy. Several techniques have been attempted, from traditional hand-engineered features to present day deep learning and state space models.

\subsection{Traditional Machine Learning Methods}

Previous methods for seizure detection relied on manual feature selection in which statistical, spectral and wavelet features were computed from the EEG signal that is used by traditional machine learning (ML) models like Support Vector Machines (SVM) \cite{liu2012automatic}, k-Nearest Neighbours (kNN) \cite{wang2020improved} and Random Forests \cite{acharya2012automated, ocak2008optimal}. Although these techniques demonstrated preliminary success, the fact that they were based on expert knowledge and handcrafted features made it difficult to generalize them to different EEG patterns as well as patients.

\subsection{Deep Learning Advances}

The paradigm shifted toward end-to-end models that can learn directly from raw or minimally preprocessed EEG signals with the advent of deep learning. Convolutional Neural Networks (CNNs) \cite{lawhern2018eegnet,schirrmeister2017deep,tang2021self,lian2020learning} have demonstrated strong performance in spatial pattern extraction from EEG signals \cite{10.3389/fneur.2024.1389731}, while Recurrent Neural Networks (RNNs) \cite{vidyaratne2016deep,talathi2017deep} and Long Short-Term Memory (LSTM) models have been widely adopted for capturing temporal dependencies \cite{10.1007/978-981-99-4742-3_25}. For instance, SlimSeizNet \cite{Lu_2025} and SOUL Online \cite{9773289} demonstrated robust performance while emphasizing model compactness and online inference, respectively. More recent models such as Spiking Conformer \cite{Chen_2024} and DLBSP \cite{https://doi.org/10.1002/eng2.12918} continue to push the boundaries of accuracy and efficiency by leveraging novel architectures. \cite{hussein2018epileptic}

\subsection{State Space Models (SSMs)}

SSMs, or structured state space models, provide a mathematical framework that is ideal for simulating temporal and sequential data.  Despite their strength, classical SSMs are frequently computationally costly and difficult to scale in real-time EEG situations.  Graph-based SSMs with spatiotemporal attention mechanisms can perform exceptionally well on seizure prediction tasks, as shown by recent advancements like SGSTAN \cite{Xiang2025}.  EEGMamba \cite{gui2024eegmambabidirectionalstatespace} achieves state-of-the-art results on benchmark datasets by combining bidirectional context modeling with Mamba-style selective recurrence to further improve SSM performance.

\subsection{The Mamba SSM Architecture}

Selective structured state space layers are introduced by Mamba, a hardware-efficient SSM framework, which drastically lowers computational requirements without sacrificing powerful sequence modeling capabilities \cite{mamba, gu2021efficiently, auger2021guide}.  Real-time applications' resource limitations are well suited to this design.  Models like ConvMambaNet can train about three times faster than similar CNN, RNN, and Transformer architectures thanks to Mamba's high throughput and low latency achieved by utilizing GPU parallelism.  For time-sensitive tasks like real-time seizure detection, where quick signal processing is just as important as precise prediction, Mamba's effective speed-accuracy ratio makes it especially appropriate.

\subsection{Recent CHB-MIT Seizure Detection Studies}

The CHB-MIT Scalp EEG Database, which provides detailed, real-world annotations of pediatric seizure episodes, has emerged as a standard for assessing seizure detection models.  On this dataset, a number of models report high sensitivity \cite{Xiang2025} and accuracy: CNN-BiGRU \cite{10.1007/978-981-99-4742-3_25} prioritizes low false-positive rates; SGSTAN achieved 98.2\% accuracy and 97.85\% sensitivity; EEGMamba achieved 98.6\% accuracy with 98.7\% F1-score\cite{gui2024eegmambabidirectionalstatespace}.  Additionally, single-channel and self-supervised learning techniques have been investigated, emphasizing the field's emphasis on accurate but lightweight solutions.

Despite these developments, the majority of earlier architectures are either constrained by their incapacity to manage the variability of EEG signals across subjects or prioritize accuracy at the expense of inference time.  Our work suggests a hybrid ConvMambaNet model that combines Mamba SSM and convolutional feature extraction for effective and precise seizure detection in order to close this gap.  On the CHB-MIT dataset, this architecture tackles important issues like class imbalance, temporal modeling, and real-time performance.

\begin{algorithm}[t]
	\SetAlgoVlined
	\KwIn{Batch size $B$, Sequence length $L$, Input dimension $d_{\text{input}}$, Model parameters: $d_{\text{model}}$, $d_{\text{state}}$, $d_{\text{conv}}$, and $e$ (expansion factor).}
    \KwOut{Output tensor $y$ with the same shape as input tensor $x$.}
	\caption{Mamba Model Baseline Algorithm} 
	\label{alg:mamba_baseline}
	\textbf{Step 1: Input Initialization} \\
	Define input tensor $x \in \mathbb{R}^{B \times L \times d_{\text{input}}}$, where $x$ is randomly initialized as:
	\[
	x \sim \mathcal{N}(0, 1)
	\]
	Load the tensor onto a CUDA device.\\
	\textbf{Step 2: Model Configuration} \\
	Initialize the Mamba model with the following parameters:
	\begin{itemize}
	    \item Model dimension: $d_{\text{model}} = d_{\text{input}} = 16$,
	    \item SSM state expansion factor: $d_{\text{state}} = 16$,
	    \item Local convolution width: $d_{\text{conv}} = 4$,
	    \item Block expansion factor: $e = 2$.
	\end{itemize}
	Load the model onto a CUDA device.\\
	\textbf{Step 3: Model Forward Pass} \\
	Pass the input tensor $x$ through the model:
	\[
	y = \text{Mamba}(x)
	\]
	\textbf{Step 4: Output Validation} \\
	Assert that the output tensor $y$ has the same shape as the input tensor $x$:
	\[
	y \in \mathbb{R}^{B \times L \times d_{\text{input}}}
	\]
	\textbf{Output:} Return the output tensor $y$.\\
\end{algorithm}

\begin{figure*}
     \centering
     \includegraphics[width=\textwidth]{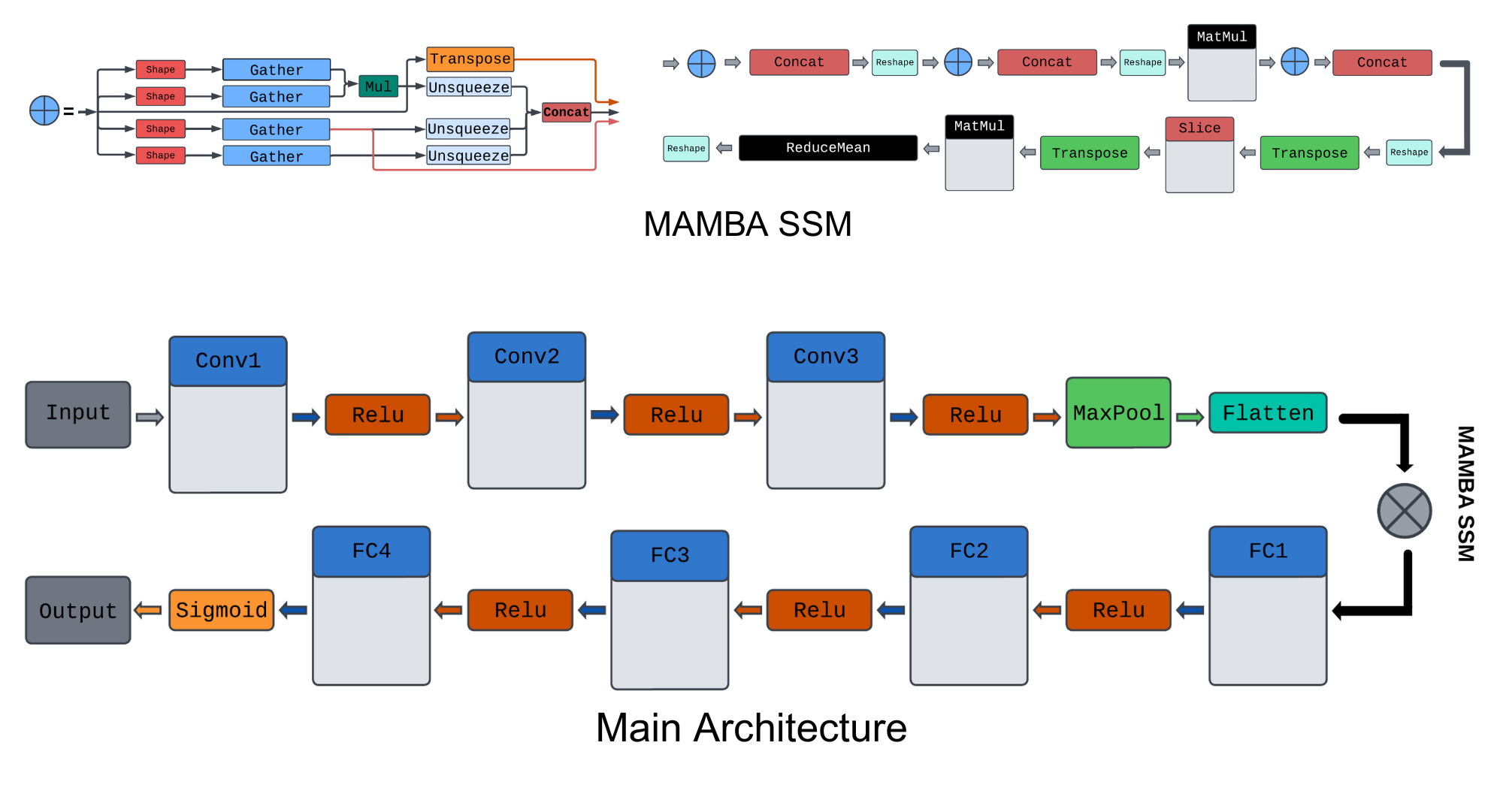}
     \caption{Overview of ConvMambaNet architecture combining CNN feature extraction with Mamba SSM for temporal modeling and classification. The Mamba SSM module employs selective state-space transformations with gating mechanisms to capture long-range sequential dependencies.}
\end{figure*}

\section{Methodology}
\label{Methodology}

We have significantly modified the model in this paper by incorporating Mamba SSM into CNN architecture. This modification was made to enable temporal analysis of complex and time-dependent EEG signals.

\subsection{Enhanced ConvMambaNet Model Integration}

The novelty of the proposed model is that it incorporates the state-of-the-art ConvMambaNet, a hybrid architecture based on Mamba State-Space Model (Mamba-SSM) and CNN for handling particular difficulties in analyzing EEG signals. The EEG signals have order dependencies and complex temporal dynamics, hence a model that captures both spatial and temporal factors is needed. ConvMambaNet is able to satisfy these constraints well and it combines the best of both Mamba-SSM and CNNs. The Mamba-SSM block is a core piece of ConvMambaNet, learning the temporal dynamics of EEG signal via state-space representation. The Mamba-SSM block is developed to model temporal patterns that are important for discriminating epileptic seizures whereas traditional CNNs do good at the spatial feature extraction and ignore the consecutive dependencies. Layer-wise initialization is utilized by ConvMambaNet to keep the training stable and efficient. Weights in convolutions are initialized at He/Kaiming with zero biases, while Batch Normalization begins with $\gamma{=}1$, $\beta{=}0$. Fully connected layer use Glorot initialization with zero bias. In the Mamba-SSM, we use a stability-oriented parameterization with $\mathbf{A}{=}-\operatorname{softplus}(\mathbf{A}_{\log})$ so that eigenvalues are non-positive at initialization; $\mathbf{B}$ and $\mathbf{C}$ are zero-mean with variance $1/\sqrt{d_{\text{state}}}$, and on the $\Delta t$ projection, there is a small negative bias and weak weights. He initialization with $d_{\text{conv}}{-}1$ padding is applied in the depthwise convolution before the selective scan. Collectively, these choices maintain the dynamical properties well-conditioned and alleviate exploding/vanishing gradients on long EEG sequences. The fully connected layers of ConvMambaNet have the dropout terms utilized to reduce overfitting \cite{lim2020study}. Such extension is especially significant because of the high between-subject variability and scarcity of annotated seizures in EEG datasets. Furthermore, multi-head attention mechanisms \cite{vaswani2017attention} are embedded into the Mamba-SSM block to enhance the ability of sequence modeling. These attention mechanisms allow the model to learn which temporal regions in the EEG signals are critical, and therefore help it better identify patterns associated with seizures.

\subsection{Hybrid Framework for EEG Analysis}

The hybrid structure of ConvMambaNet integrates the spatial characteristic and representation capacity from CNN layers, as well as the temporal modeling property from Mamba-SSM block. Such combination indeed provides an in-depth feature extraction on EEG data to cope with the two-dimensional complexity of spatial and sequential dependencies. Convolutional layers capture the high level spatial information of raw EEG signals while the Mamba-SSM block processes these features in time. This multicentric approach also gives an overview that is necessary for a meaningful seizure detection. ConvMambaNet is completely end-to-end trainable and adopts the Mamba-SSM block into the CNN architecture, trained by Adam optimizer \cite{ravichandranadam}. This approach eliminates the necessity of feature construction by humans, making it applicable to various EEG datasets and clinical use. The training pipeline involves several preprocessing steps like balanced sampling of seizure and non-seizure events which make the model sensitive to rare seizure events, but specific to non-seizure events.

\section{Experiments}
\label{Experiments}
\subsection{Dataset}
\label{Dataset}

We conducted our experiments with the CHB-MIT Scalp EEG Dataset \cite{goldberger2000physiobank}, a popular benchmark in epilepsy community. It includes EEGs from 22 sedated pediatric patients with refractory epilepsy acquired at the Children’s Hospital Boston. The dataset contains 664 EEG files which have recording time that ranges from about 1 to 4 hours. The samples are actually digitized at 256 Hz and 16-bits. A total of 198 seizure episodes (182 from 23 subjects) are labelled with exact onset-offset timing, allowing the effective training and testing for a model performing both seizure detection and classification.

\begin{figure}[h!]
\noindent 
\begin{minipage}{0.5\linewidth} 
    \includegraphics[width=\linewidth]{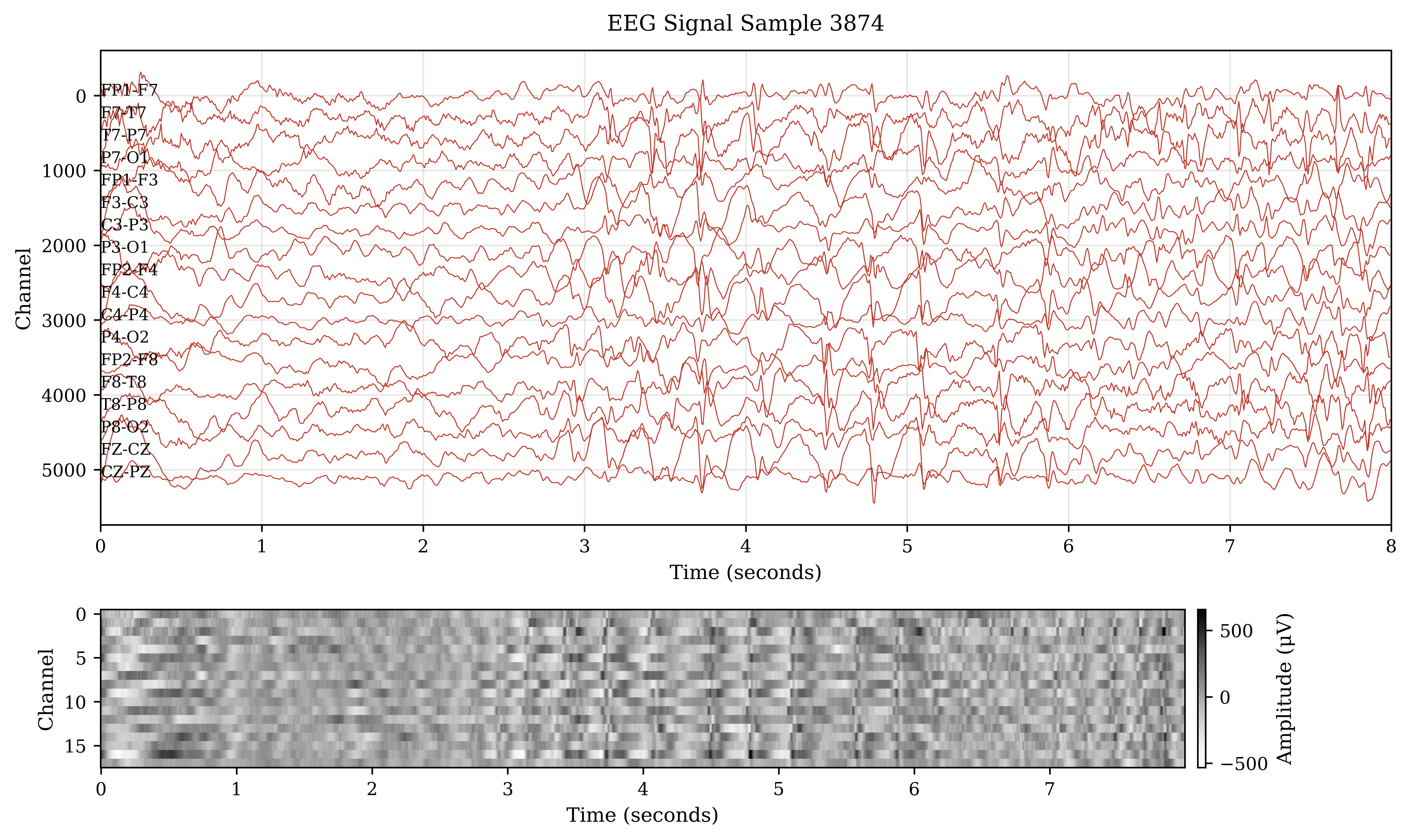}\\[2pt]
    \includegraphics[width=\linewidth]{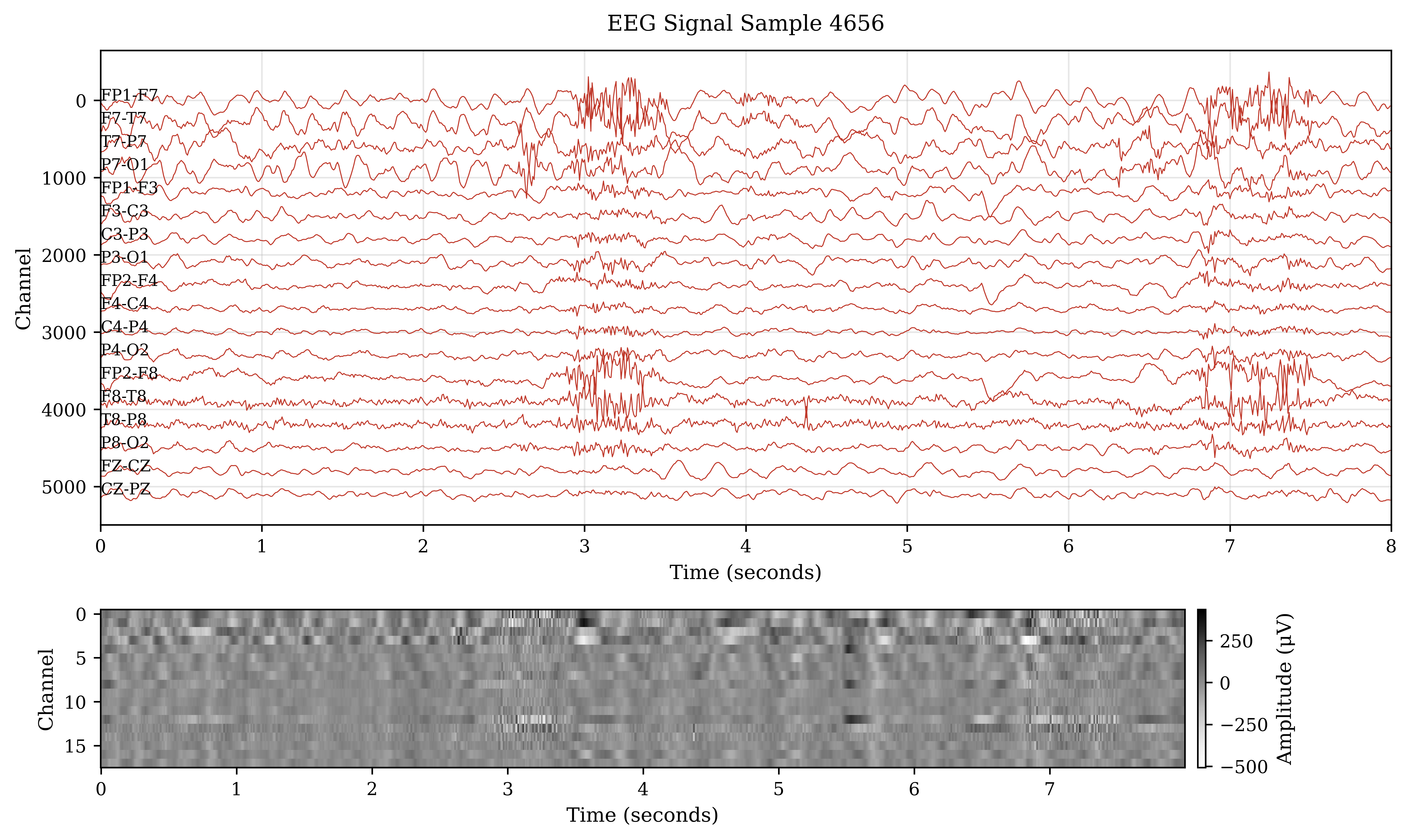}\\[2pt]
    \includegraphics[width=\linewidth]{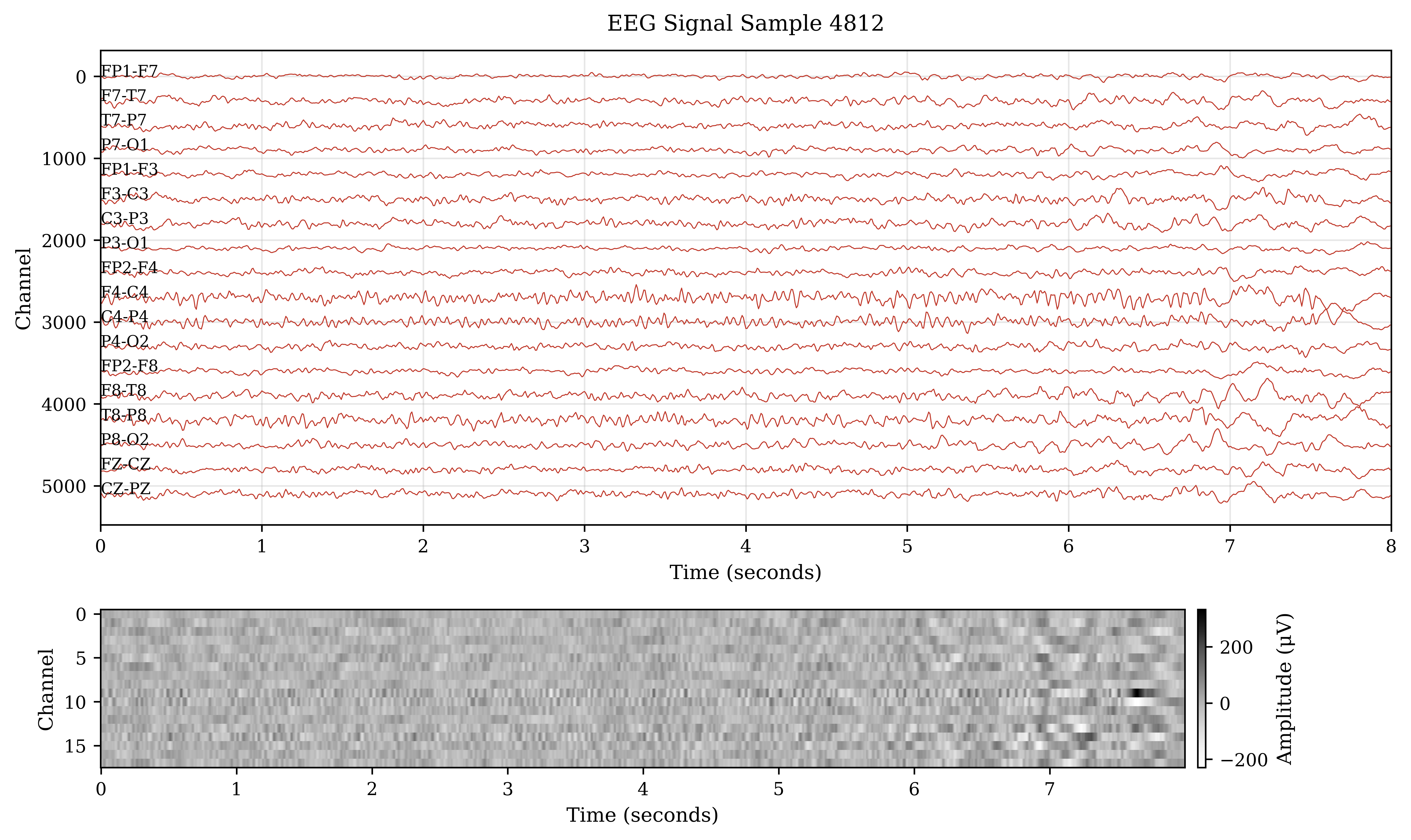}\\[2pt]
    \includegraphics[width=\linewidth]{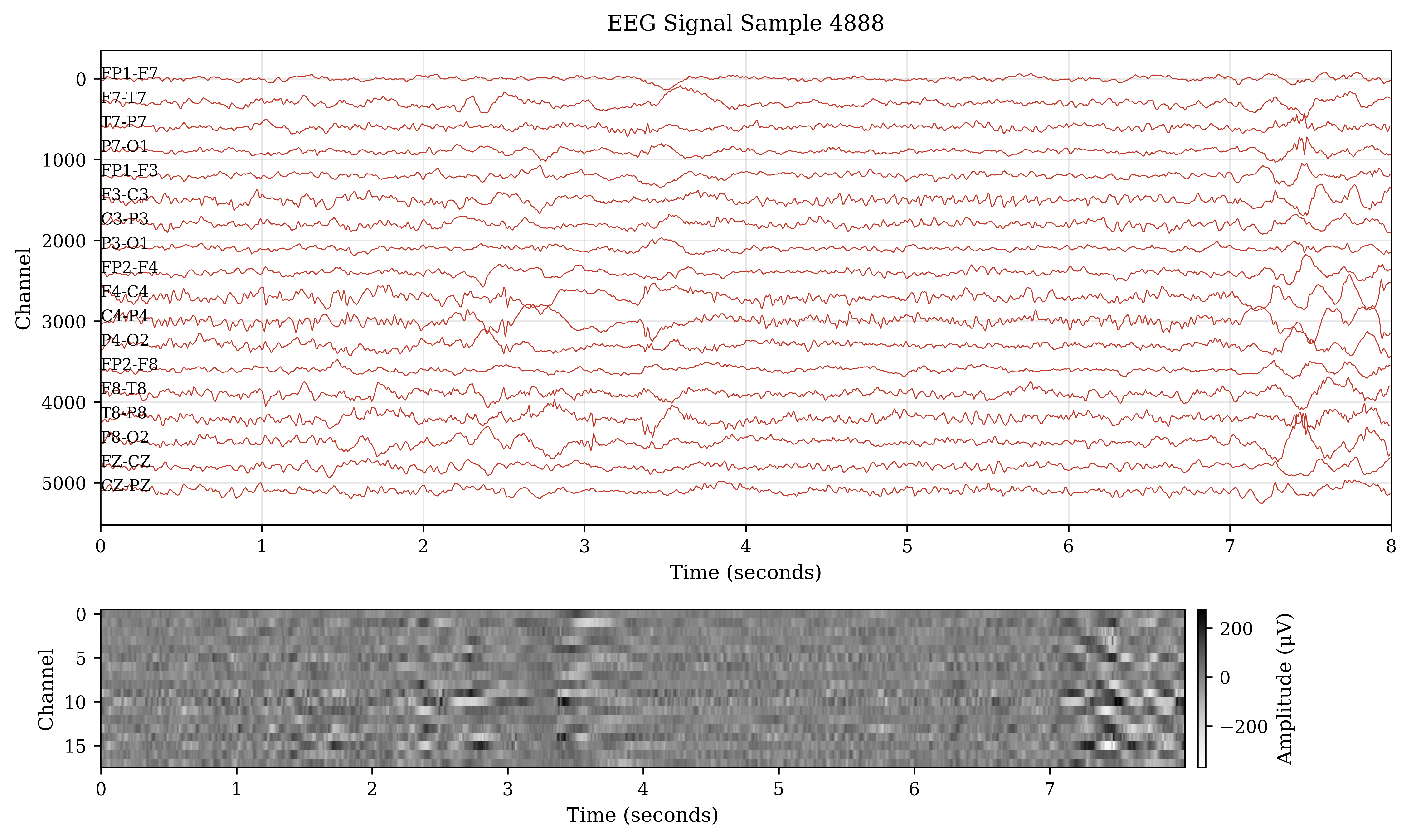}
\end{minipage}
\hfill
\begin{minipage}{0.48\linewidth}
    \includegraphics[width=\linewidth]{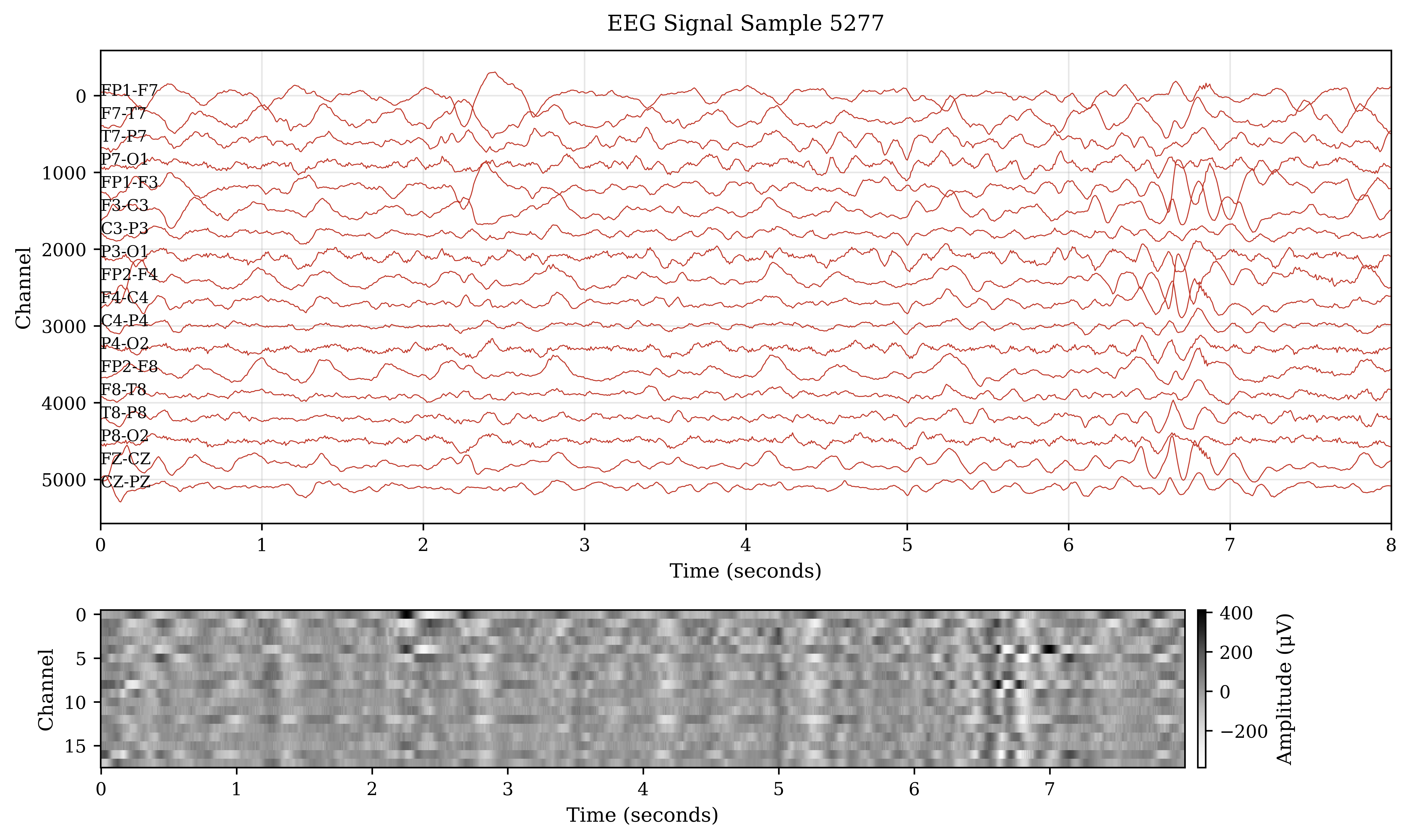}\\[2pt]
    \includegraphics[width=\linewidth]{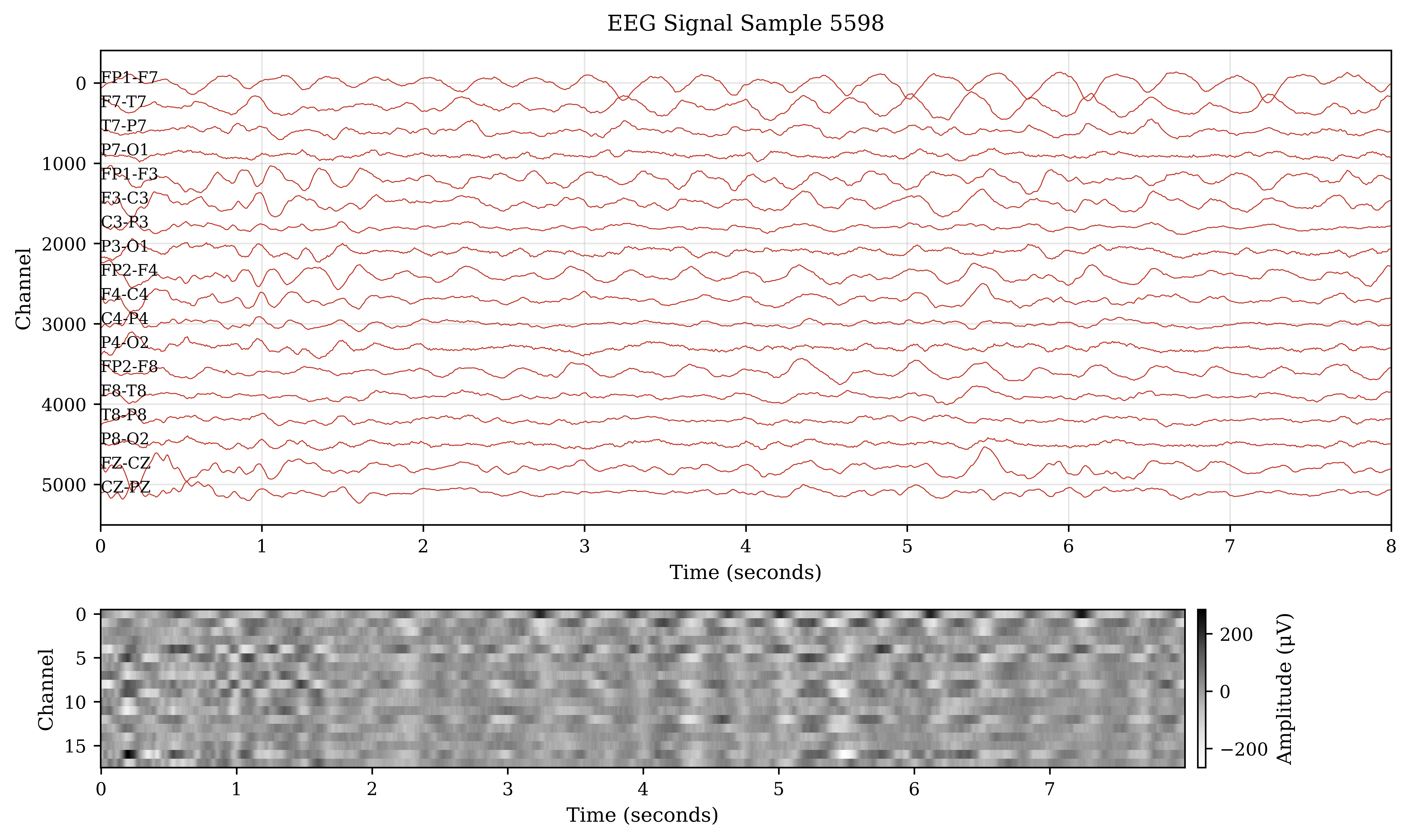}\\[2pt]
    \includegraphics[width=\linewidth]{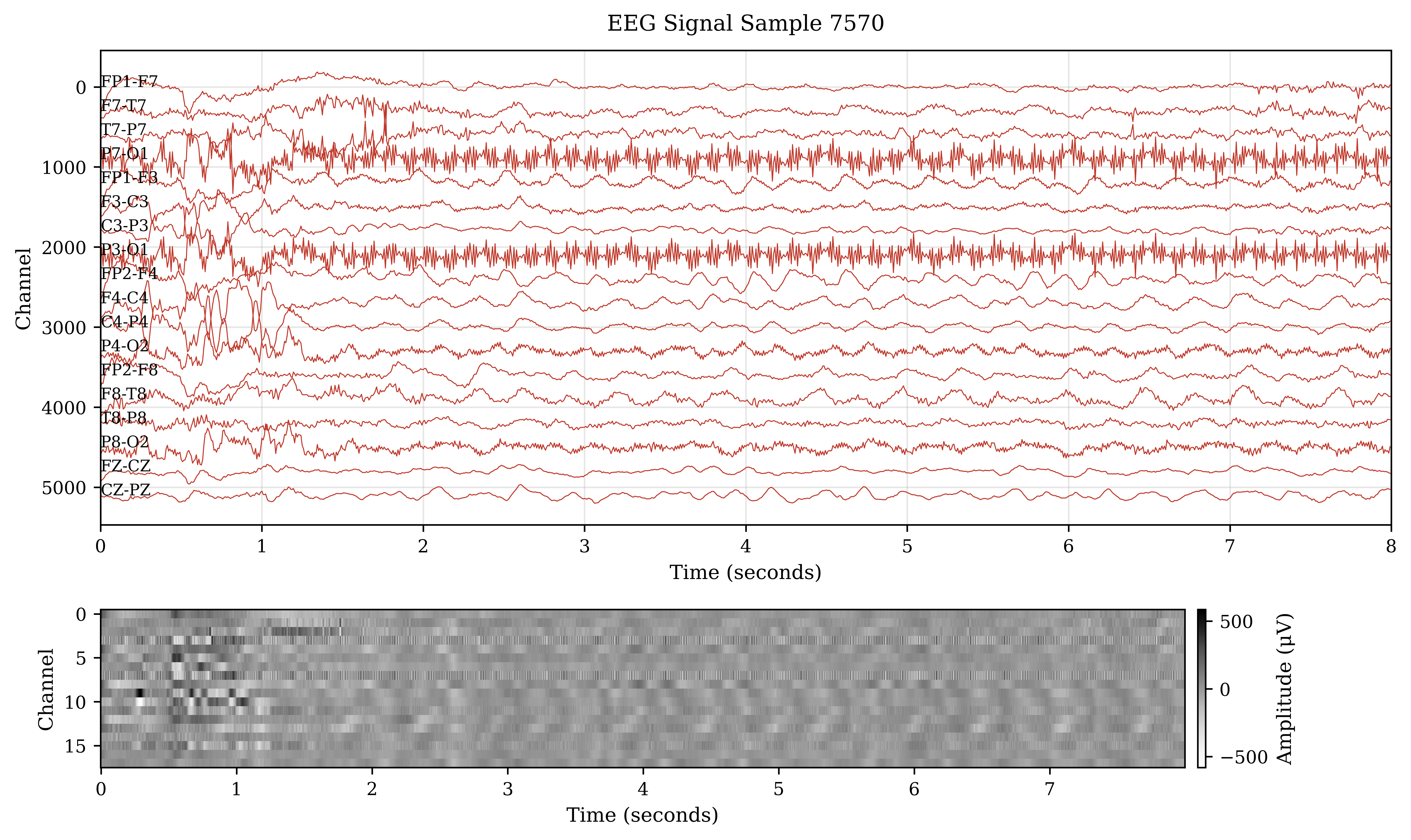}\\[2pt]
    \includegraphics[width=\linewidth]{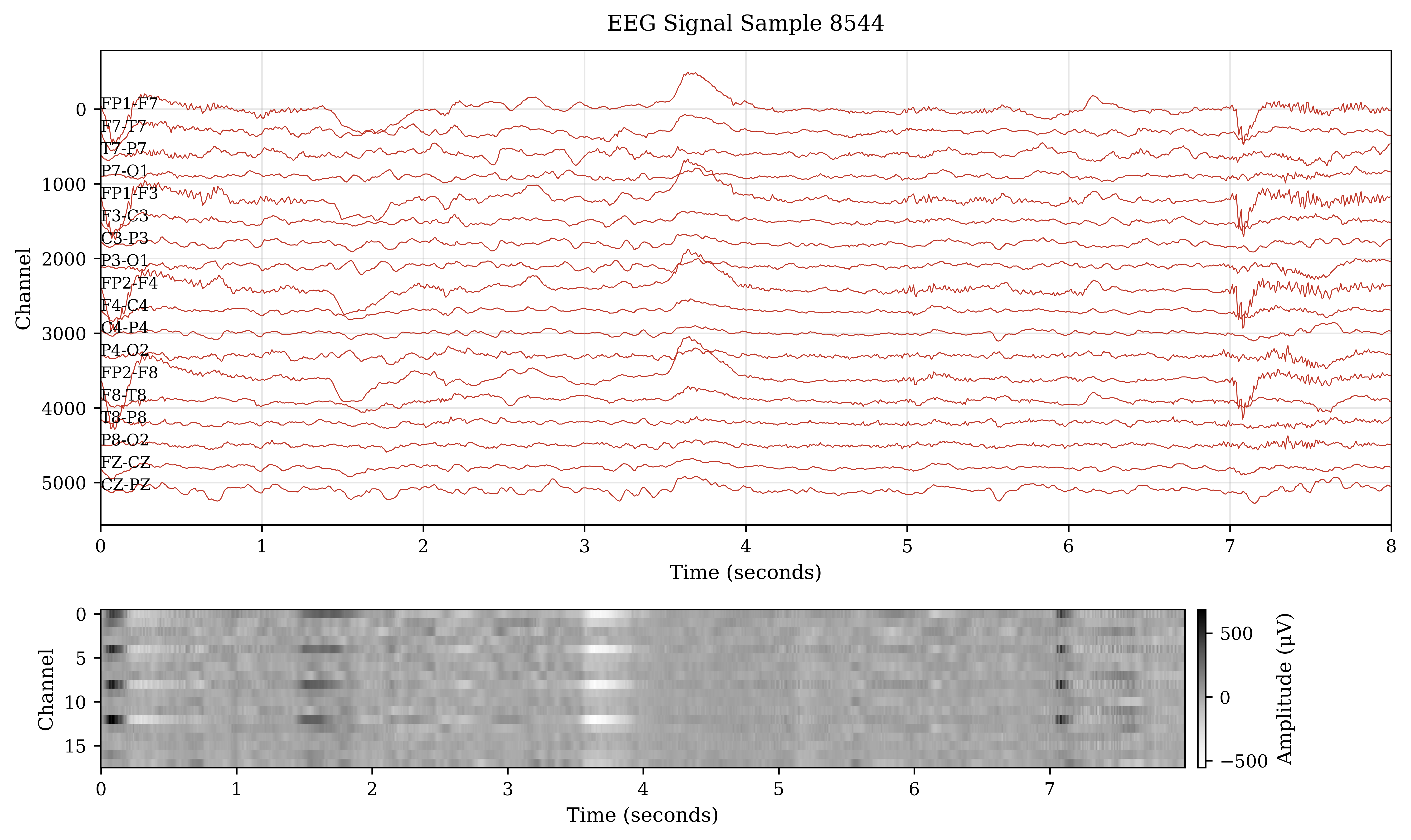}
\end{minipage}
\caption{EEG Signals with Multiple Channels}
\label{fig:eeg_signals}
\end{figure}

\subsection{Data Preprocessing}
\label{Data_Preprocessing}
\definecolor{deeporange}{HTML}{FFCC80}
Cleaning the raw EEG data is an essential stage to guarantee the quality and validity of the analysis presented here. This work used the CHB-MIT Scalp EEG Database, which contains several patients and the annotated seizure events. What follows is the preprocessing workflow carried out on database. Channel Selection and Mapping: The bipolar EEG channels in Table \ref{tab:channel_labels} were chosen as a common array of channel names, centering around the labels that were qualitatively found to be consistent across subjects in the training set (CHB-MIT) based on clinical relevance:

\begin{minipage}{\linewidth}
\begin{table}[H]
\centering
\renewcommand{\arraystretch}{1.7}
\setlength{\tabcolsep}{5.4pt}
\begin{tabular}{|c|c|c|c|c|c|}
\hline
\rowcolor{deeporange}
    \texttt{FP1-F7} & \texttt{F7-T7} & \texttt{T7-P7} & \texttt{P7-O1} & \texttt{FP1-F3} & \texttt{F3-C3} \\
\hline
\rowcolor{deeporange}
    \texttt{C3-P3} & \texttt{P3-O1} & \texttt{FP2-F4} & \texttt{F4-C4} & \texttt{C4-P4} & \texttt{P4-O2} \\
\hline
\rowcolor{deeporange}
    \texttt{FP2-F8} & \texttt{F8-T8} & \texttt{T8-P8} & \texttt{P8-O2} & \texttt{FZ-CZ} & \texttt{CZ-PZ} \\
\hline
\end{tabular}
\caption{Array of selected EEG bipolar channels used in preprocessing.}
\label{tab:channel_labels}
\end{table}
\end{minipage}

Preprocessing started with scanning through the dataset to find and organize patient folders. Patients were randomly selected, with 80\% of the dataset used for training and 20\% for testing. EEG data of each patient was taken and processed by means of the MNE library \cite{esch2019mne, samarpita2023differentiating}. Acquisition settings including sample rate and epoch were verified. Special care was taken in parsing the unique channel names. Seizure annotation: In order to facilitate the seizure detection process, manual annotations of the seizure events were produced with the help of WFDB (Waveform Database) library \cite{wfdb2023}. A binary timecourse representing moments of seizure was produced (at timepoints where there was a seizure, 1; otherwise, 0). This labelling was designed to capture the seizures as they appeared in the data. Signals were divided into subsets of overlapping time windows in order to generate adequate samples for model training. One comparison window was 8 seconds long with a step size of 4 seconds and therefore ensured close to complete sampling of the EGG data. The size of the window was conditioned to the sampling rate of the EEG recordings. Processed signal samples and corresponding seizure labels were stored as NumPy \cite{harris2020array} arrays, for faster access during model training and evaluation. The final dataset comprised 9,505 compiled signals, where 2,581 were diagnosed to contain a seizure. The dataset was divided into training and test sets via stratified sampling. Using this method, the ratio of seizure cases and non-seizure cases is kept at its original value in training as well as validation datasets. We reduced the potential for bias by stratifying the split based on seizure labels and encourage models to generalize across classes. We also ensure there is no overlap in the recordings or samples between the two subsets, so that the model can learn from a strong and fair evaluation during training.

\subsection{Evaluation Metrics}
\label{Evaluation_Metrics}

The effectiveness of our EEG seizure detection model was evaluated using a number of common classification metrics.  Precision, recall, accuracy, F1-score, and the area under the receiver operating characteristic curve (AUC-ROC) are all included.  Combining these metrics provides an overall picture of the model's sensitivity/specificity performance, general classification accuracy at various thresholds, and ability to distinguish between seizure and non-seizure events.

\section{Numerical Analysis}
\label{Numerical_Analysis}
\subsection{Results}
\label{Results}

The ConvMambaNet model also demonstrated good performance on the CHB-MIT EEG dataset, with precision, recall, and F1-score of 1.00, 0.99, and 0.99 for all tested samples.  The weighted F1-score was 0.99 and the final classification accuracy was 99\%.  These results show how well the model performs in noisy clinical data and show how useful it is for differentiating EEG events.

\begin{figure}[htbp]
    \centering
    \begin{minipage}{\linewidth}
        \centering
        \includegraphics[width=\linewidth]{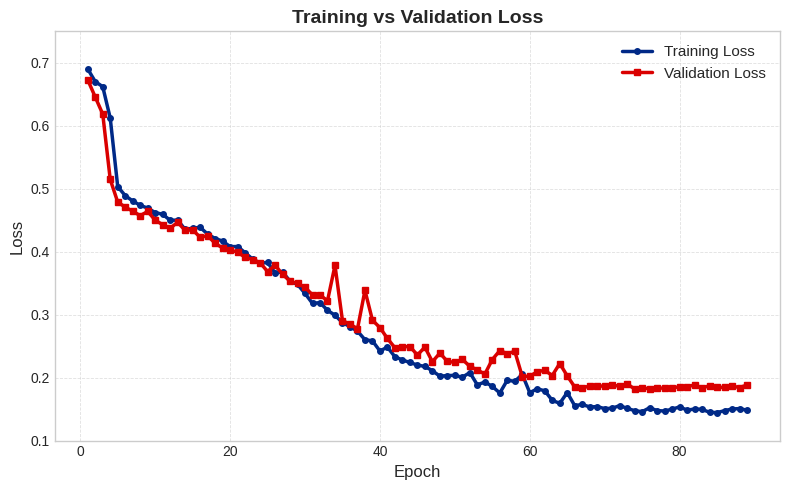}
        \caption{Training and Validation Loss}
        \label{fig:loss_curve}
    \end{minipage}\vfill
    \begin{minipage}{\linewidth}
        \centering
        \includegraphics[width=\linewidth]{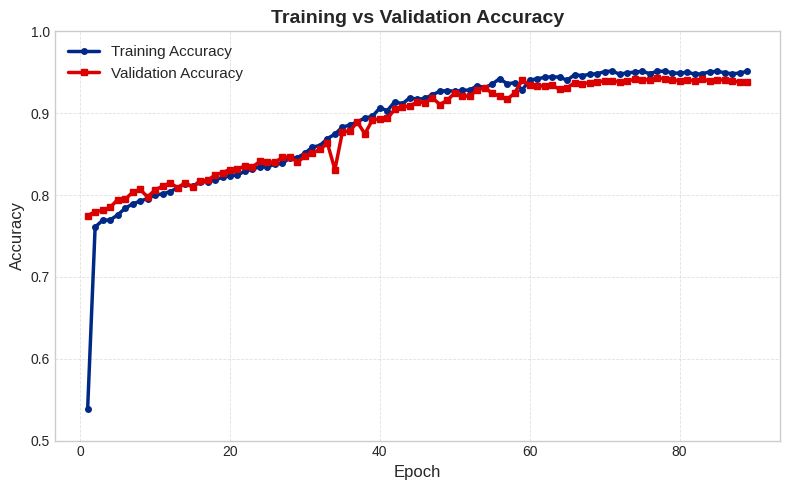}
        \caption{Training and Validation Accuracy}
        \label{fig:accuracy_curve}
    \end{minipage}
\end{figure}

\begin{figure}[htbp]
    \centering

    \subfloat[Confusion Matrix of ConvMambaNet Architecture]{%
        \includegraphics[width=0.8\linewidth]{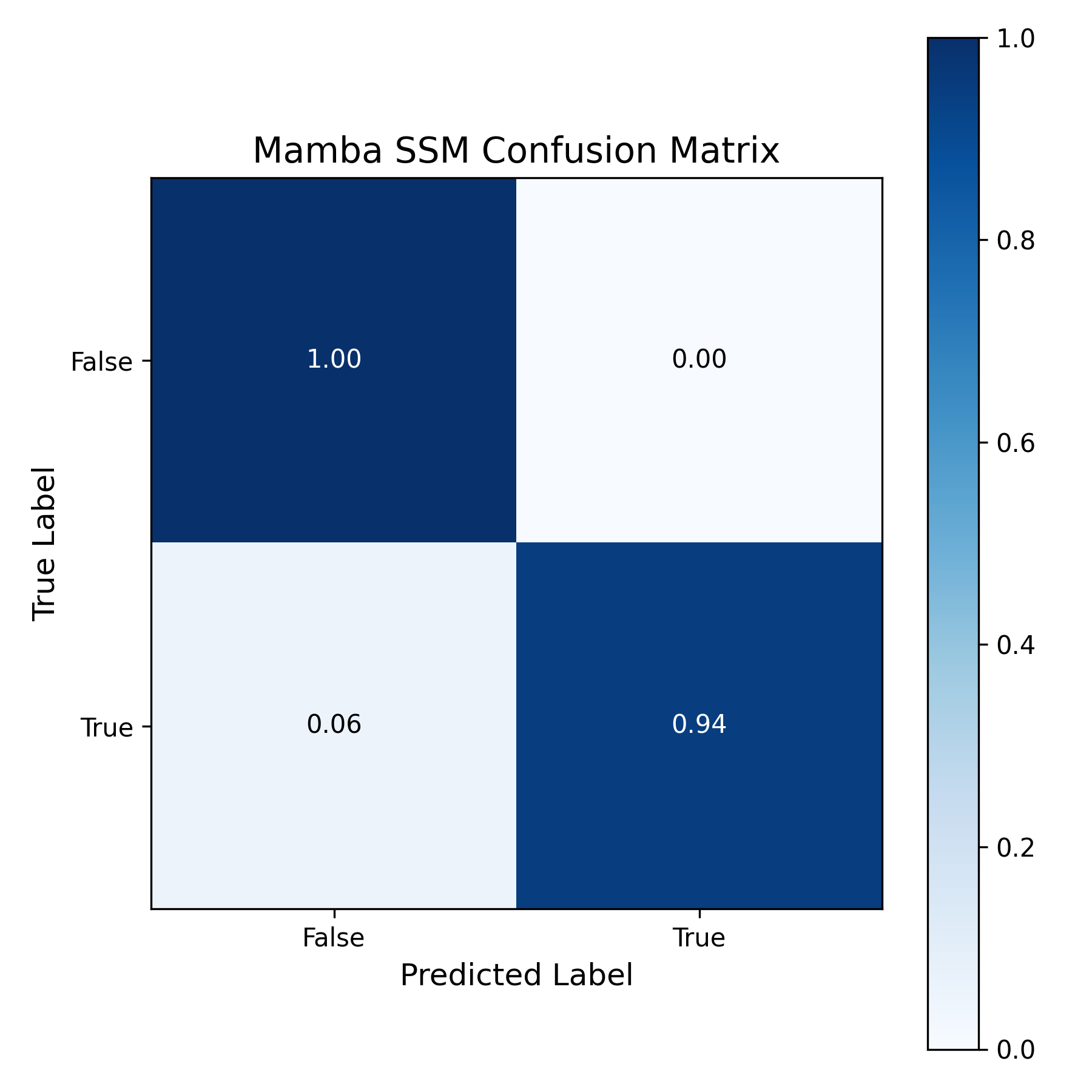}%
    }

    \vspace{1.5em} 

    \hspace{-1cm}
    \subfloat[ROC Curve AUC Score of ConvMambaNet Architecture]{%
        \includegraphics[width=0.8\linewidth]{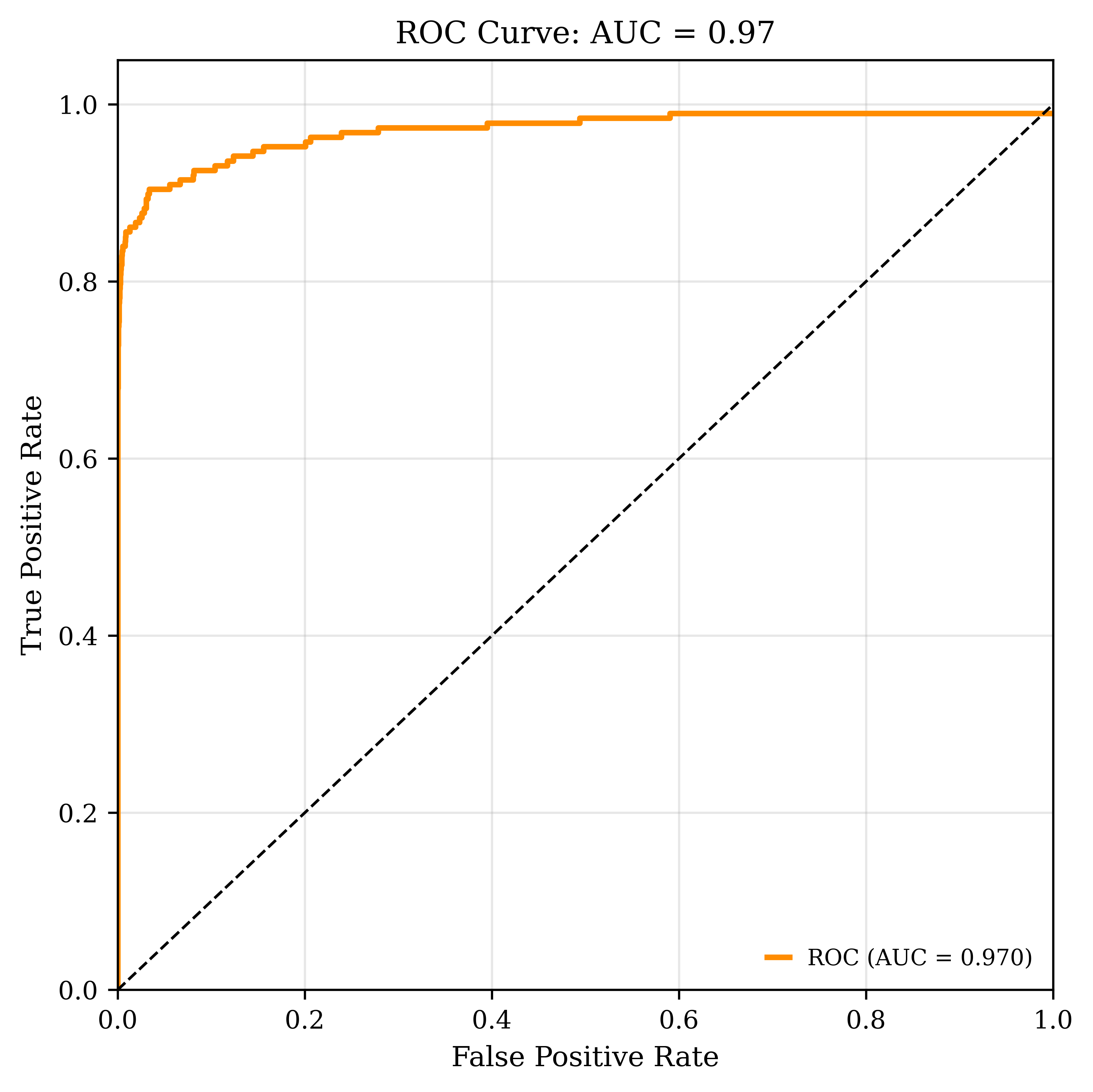}%
    }

    \caption{(a) Confusion matrix showing class-level prediction accuracy, and (b) ROC curve illustrating the model's discriminative power.}
    \label{fig:confusion-roc-vertical}
\end{figure}

The training and validation curves regularly converge without overfitting, as seen in Figures Fig. \ref{fig:loss_curve} and \ref{fig:accuracy_curve}.  In the EEG-based seizure-detection task, ConvMambaNet based on Mamba-SSM performed exceptionally well in detecting non-seizure (False) events.  With an accuracy of 99\%, the model achieved nearly perfect precision (0.99), recall, and F1-score (0.99, 0.99) for the False class.  The model can distinguish between seizure and non-seizure events, as evidenced by its AUC of 0.97 (ROC curve in Fig. \ref{fig:confusion-roc-vertical}).  Additionally, ConvMambaNet's high specificity and good sensitivity are validated by the confusion matrix shown in Figure \ref{fig:confusion-roc-vertical}.  These experimental findings confirm the model's resilience in EEG signal classification. These results emphasize the model's excellent ability to process non-seizure based patterns, which is really important for reducing false positives in clinical real-time use.

\subsection{Comparison with Other Models}

We conducted comparisons of the ConvMambaNet model with the state-of-the-art models including RNN based and CNN-based ones as well as Transformer models. In Table ~\ref{tab:seizure_comparison}, we can see that ConvMambaNet achieves the highest precision (0.99), recall (0.99), F1-score (0.99), AUC(0.97) and accuracy rates(99\%) for seizure detection as well as non-seizure detection than traditional deep learning baselines by a significant margin.

ConvMambaNet significantly outperforms existing models in the literature for EEG-based seizure detection in both accuracy and robustness. In this problem the SGSTAN \cite{Xiang2025} model had exactly the same performance as our method with F1=97 and accuracy= 98\%, which supports its achievements regarding sequential graph based state space modeling. The bidirectional state-space learning EEGMamba \cite{gui2024eegmambabidirectionalstatespace} had F1-score 0.79 and AUC 0.91 making gains in including the temporal context, however it still can be further optimized. On the other hand, the compactness and real-time property of the model was emphasized in SlimSeizNet \cite{Lu_2025}, reaching recall = 95\%, accuracy = 94\%. Single-channel deep learning methods \cite{10.3389/fneur.2024.1389731} obtained very high recall (0.97) and may be applicable when few channels are available.

Spiking Conformers \cite{Chen_2024} introduced successful spiking neural network dynamics while achieving 94\% recall and 97\% accuracy that bridges the gap of quality between biological realism and deep learning performance. Self-supervised transformer models \cite{huang2025self} presented an alternative path to unsupervised representation learning yet learned seizure detection with significantly lower F1-score (0.59) than ours, and also with diminished accuracy (80\%), hence challenging the possibility of purely transformer-based solutions for seizure detection. Hybrid attended CNN-BiGRU \cite{10.1007/978-981-99-4742-3_25} and BRR maintained the above 93\% recall, and achieved above 98\% of accuracy when assessing network capacity for combining spatial with temporal modeling. The efficiency of online inference is the primary consideration for SOUL Online \cite{9773289} by obtaining an F1 recall and accuracy of 97\% and 95\%, respectively based on clinical deployment demands. The recall of the DLBSP method \cite{https://doi.org/10.1002/eng2.12918} reached 90\%, which indicates that deep learning-based seizure prediction becomes much more mature.

On the other hand, ConvMambaNet exhibits superior precision, recall, F1-score, AUC and overall accuracy to the above models in all cases which indicates its capability of capturing long range temporal dependencies as well as fine-grained EEG features essential for robust seizure detection. Given both these aspects, this relatively absolute performance gain makes ConvMambaNet a desirable candidate for practical clinical use.

\begin{table}[htb!]
\centering
\renewcommand{\arraystretch}{1.4}
\setlength{\tabcolsep}{4.3pt}
\begin{NiceTabular}{lcccccc}[color-inside]
\CodeBefore
 \rowlistcolors{2}{,tab2}[restart,cols={1-6}]
\Body 
\arrayrulecolor{black}\toprule
\cellcolor{red!25}\textbf{Model}  & \cellcolor{red!25}\textbf{Precision} & \cellcolor{red!25}\textbf{Recall} & \cellcolor{red!25}\textbf{F1-Score} & \cellcolor{red!25}\textbf{AUC} & \cellcolor{red!25}\textbf{ACC} \\

\arrayrulecolor{black}\midrule
\cellcolor{green!10}CNN & \cellcolor{green!10}0.96 & \cellcolor{green!10}0.93 & \cellcolor{green!10}0.94 & \cellcolor{green!10}0.92 & \cellcolor{green!10}96\% \\
\cellcolor{green!15}RNN & \cellcolor{green!15}0.94 & \cellcolor{green!15}0.90 & \cellcolor{green!15}0.92 & \cellcolor{green!15}0.89 & \cellcolor{green!15}92\% \\
\cellcolor{green!10}Transformer & \cellcolor{green!10}0.98 & \cellcolor{green!10}0.95 & \cellcolor{green!10}0.97 & \cellcolor{green!10}0.95 & \cellcolor{green!10}94\% \\

\cellcolor{green!30}\textbf{ConvMambaNet*} & \cellcolor{green!30}\textbf{0.99*} & \cellcolor{green!30}\textbf{0.99*} & \cellcolor{green!30}\textbf{0.99*} & \cellcolor{green!30}\textbf{0.97*} & \cellcolor{green!30}\textbf{99\%*} \\
\cellcolor{blue!5}SGSTAN \cite{Xiang2025} & \cellcolor{blue!5}— & \cellcolor{blue!5}0.97 & \cellcolor{blue!5}0.97 & \cellcolor{blue!5}— & \cellcolor{blue!5}98\% \\
\cellcolor{blue!10}EEGMamba \cite{gui2024eegmambabidirectionalstatespace} & \cellcolor{blue!10}— & \cellcolor{blue!10}— & \cellcolor{blue!10}0.79 & \cellcolor{blue!10}0.91 & \cellcolor{blue!10}97\% \\
\cellcolor{blue!5}SlimSeizNet \cite{Lu_2025} & \cellcolor{blue!5}— & \cellcolor{blue!5}0.95 & \cellcolor{blue!5}— & \cellcolor{blue!5}— & \cellcolor{blue!5}94\% \\
\cellcolor{blue!10}Single-Channel DL \cite{10.3389/fneur.2024.1389731} & \cellcolor{blue!10}— & \cellcolor{blue!10}0.97 & \cellcolor{blue!10}— & \cellcolor{blue!10}— & \cellcolor{blue!10}— \\
\cellcolor{blue!5}Spiking Conformer \cite{Chen_2024} & \cellcolor{blue!5}— & \cellcolor{blue!5}0.94 & \cellcolor{blue!5}— & \cellcolor{blue!5}— & \cellcolor{blue!5}97\% \\
\cellcolor{blue!10}\makecell[l]{Self-Supervised\\Transformer \cite{huang2025self}}
 & \cellcolor{blue!10}— & \cellcolor{blue!10}— & \cellcolor{blue!10}0.59 & \cellcolor{blue!10}— & \cellcolor{blue!10}80\% \\
\cellcolor{blue!5}CNN-BiGRU \cite{10.1007/978-981-99-4742-3_25} & \cellcolor{blue!5}— & \cellcolor{blue!5}93 & \cellcolor{blue!5}— & \cellcolor{blue!5}— & \cellcolor{blue!5}98\% \\
\cellcolor{blue!10}SOUL Online \cite{9773289} & \cellcolor{blue!10}— & \cellcolor{blue!10}0.97 & \cellcolor{blue!10}— & \cellcolor{blue!10}— & \cellcolor{blue!10}95\% \\
\cellcolor{blue!5}DLBSP \cite{https://doi.org/10.1002/eng2.12918} & \cellcolor{blue!5}— & \cellcolor{blue!5}0.90 & \cellcolor{blue!5}— & \cellcolor{blue!5}— & \cellcolor{blue!5}— \\

\arrayrulecolor{black}\bottomrule
\end{NiceTabular}
\caption{Performance comparison of seizure detection models on the CHB-MIT Scalp EEG dataset. Green-highlighted rows represent our proposed hybrid ConvMambaNet model and baseline variants of CNN, RNN and Transformer in same pipeline. All metrics are rounded to 2 decimal places.}
\label{tab:seizure_comparison}
\end{table}

\section{Computational Time}

We evaluated the training efficiency of ConvMambaNet on a system with the following specifications:
\begin{itemize}
    \item CPU: AMD Ryzen Threadripper 1920X (24 threads) @ 3.50GHz
    \item GPU: NVIDIA GeForce RTX 2070 Super
    \item RAM: 32 GB
    \item Environment: Python 3.10 with TensorFlow and PyTorch
\end{itemize}

ConvMambaNet was trained with an averaged time of 48 seconds per epoch, showing superior computational efficiency compared to the baseline models. CNN needed more than 2 minutes per epoch, RNN approximately 1 minute 43 seconds and Transformers lasted for 4 minutes. With fewer training time, ConvMambaNet consistently achieved better accuracy, precision and recall, suggesting promisingness for real-time application. This trade-off between computation speed and performance, however makes ConvMambaNet an ideal solution for real-time applications like EEG-based seizure detection.

\section{Discussion}
\label{Discussion}

The presented ConvMambaNet architecture achieves superior performance than classical models (CNNs, RNNs and Transformers) in seizure detection from EEG signals. ConvMambaNet was highly robust for clinical applications, achieving 99\% accuracy and near-perfect precision, recall, and F1-scores on the non-seizure (False) class with AUC 0.97. ConvMambaNet has an advantage to model long-range dependencies in the EEG data which is needed for seizure detection. It feels like even the RNN could not have solved these types of temporal dynamic problems that CNN had faced and ConvMambaNet performs even better in this regard. It allows the stronger temporal modeling by avoiding very costly compute like Transformers while still being efficient in terms of both accuracy and FLOPs. One of ConvMambaNet’s key aspect is its computational efficiency. It has linear scaling w.r.t.\ Transformers due to quadratic sequence length reduction. OutputBack to Top ConvMambaNet adopts a hardware-sparing architecture, and thus possesses a lower time complexity and can be implemented with less memory than its formers. As a result, real-time processing of big data can be performed as in clinical seizure detection with low latency. The scalability of the method enables use with large datasets and improved generalization in various patient populations. The extremely high precision and recall for the False class implies that it can be useful to alleviate false-positives, which are crucial in medical settings. In short, ConvMambaNet strikes a good balance between performance, efficiency and scalability for real-time EEG-based seizure detection in clinical settings.

\section{Conclusion}
\label{Conclusion}

In this paper, we proposed the ConvMambaNet architecture to perform seizure detection from EEG and showed that it outperforms classical models like CNNs, RNNs and Transformers. Models:startneighborhoodplotAccording to table II, ConvMambaNet with the Mamba SSM Block added achieved an impressive accuracy of 99\% and exhibited excellent precision, recall and F1 scores for the non-seizureclassand an AUC value of0.97. These findings also demonstrate the effectiveness of the model in real-time processing EGG, which is characterized by high accuracy and computationally efficient.

The main advantage of ConvMambaNet is that it can capture long-range dependencies in EEG signals, but still depends on relatively low computation cost. Its selective state-space design enables effective sequential modeling, thus avoiding a weakness of RNN and CNN model while mitigating the computation inefficiency of Trans(Former). For this reason, ConvMambaNet is particularly suitable for clinical scenarios that demanding real-time processing, like ICU monitoring.

To conclude, ConvMambaNet offers a prospective solution for EEG seizure detection with highly accurate and time efficient as well as scalable results. That it can be transferred among patient populations and deal with a large set of data make it an advanced tool for detection of seizures in clinical services.

\section*{Authors Contribution}
\label{Authors Contribution}
\noindent
Md. Nishan Khan: Conceptualization, Data Curation, Methodology, Software, Validation, Visualization, Investigation, Writing -- Original Draft, Writing -- Review and Editing, Formal Analysis.\\
Kazi Shahriar Sanjid: Resources, Data Curation, Visualization, Validation, Writing -- Review and Editing, Investigation.\\
Md. Tanzim Hossain: Methodology, Software, Validation, Investigation.\\
Asib Mostakim Fony: Review and Editing, Resources, Visualization \\
Istiak Ahmed: Visualization, Investigation, Writing -- Review and Editing.\\
M. Monir Uddin: Supervision, Formal Analysis, Investigation, Validation, Writing -- Review and Editing, Project Administration, Funding Acquisition.

\section*{Acknowledgements}

This research was funded by the North South University Conference and Travel Grant Committee, Bangladesh (ID: CTRG-24-SEPS-20). The authors gratefully acknowledge this support, which was instrumental in the successful completion of this work

\bibliography{bibliography}

\begin{thebibliography}{10}
\providecommand{\url}[1]{#1}
\csname url@samestyle\endcsname
\providecommand{\newblock}{\relax}
\providecommand{\bibinfo}[2]{#2}
\providecommand{\BIBentrySTDinterwordspacing}{\spaceskip=0pt\relax}
\providecommand{\BIBentryALTinterwordstretchfactor}{4}
\providecommand{\BIBentryALTinterwordspacing}{\spaceskip=\fontdimen2\font plus
\BIBentryALTinterwordstretchfactor\fontdimen3\font minus \fontdimen4\font\relax}
\providecommand{\BIBforeignlanguage}[2]{{%
\expandafter\ifx\csname l@#1\endcsname\relax
\typeout{** WARNING: IEEEtran.bst: No hyphenation pattern has been}%
\typeout{** loaded for the language `#1'. Using the pattern for}%
\typeout{** the default language instead.}%
\else
\language=\csname l@#1\endcsname
\fi
#2}}
\providecommand{\BIBdecl}{\relax}
\BIBdecl

\bibitem{10.5555/3104322.3104446}
A.~Shoeb and J.~Guttag, ``Application of machine learning to epileptic seizure detection,'' in \emph{Proceedings of the 27th International Conference on International Conference on Machine Learning}, ser. ICML'10.\hskip 1em plus 0.5em minus 0.4em\relax Madison, WI, USA: Omnipress, 2010, p. 975–982.

\bibitem{schirrmeister2017deep}
R.~T. Schirrmeister, J.~T. Springenberg, L.~D.~J. Fiederer, M.~Glasstetter, K.~Eggensperger, M.~Tangermann, F.~Hutter, W.~Burgard, and T.~Ball, ``Deep learning with convolutional neural networks for eeg decoding and visualization,'' \emph{Human brain mapping}, vol.~38, no.~11, pp. 5391--5420, 2017.

\bibitem{roy2019deep}
Y.~Roy, H.~Banville, I.~Albuquerque, A.~Gramfort, T.~H. Falk, and J.~Faubert, ``Deep learning-based electroencephalography analysis: a systematic review,'' \emph{Journal of neural engineering}, vol.~16, no.~5, p. 051001, 2019.

\bibitem{liu2012automatic}
Y.~Liu, W.~Zhou, Q.~Yuan, and S.~Chen, ``Automatic seizure detection using wavelet transform and svm in long-term intracranial eeg,'' \emph{IEEE transactions on neural systems and rehabilitation engineering}, vol.~20, no.~6, pp. 749--755, 2012.

\bibitem{wang2020improved}
Z.~Wang, J.~Na, and B.~Zheng, ``An improved knn classifier for epilepsy diagnosis,'' \emph{IEEE Access}, vol.~8, pp. 100\,022--100\,030, 2020.

\bibitem{acharya2012automated}
U.~R. Acharya, F.~Molinari, S.~V. Sree, S.~Chattopadhyay, K.-H. Ng, and J.~S. Suri, ``Automated diagnosis of epileptic eeg using entropies,'' \emph{Biomedical signal processing and control}, vol.~7, no.~4, pp. 401--408, 2012.

\bibitem{ocak2008optimal}
H.~Ocak, ``Optimal classification of epileptic seizures in eeg using wavelet analysis and genetic algorithm,'' \emph{Signal processing}, vol.~88, no.~7, pp. 1858--1867, 2008.

\bibitem{lawhern2018eegnet}
V.~J. Lawhern, A.~J. Solon, N.~R. Waytowich, S.~M. Gordon, C.~P. Hung, and B.~J. Lance, ``Eegnet: a compact convolutional neural network for eeg-based brain--computer interfaces,'' \emph{Journal of neural engineering}, vol.~15, no.~5, p. 056013, 2018.

\bibitem{tang2021self}
S.~Tang, J.~A. Dunnmon, K.~Saab, X.~Zhang, Q.~Huang, F.~Dubost, D.~L. Rubin, and C.~Lee-Messer, ``Self-supervised graph neural networks for improved electroencephalographic seizure analysis,'' \emph{arXiv preprint arXiv:2104.08336}, 2021.

\bibitem{lian2020learning}
Q.~Lian, Y.~Qi, G.~Pan, and Y.~Wang, ``Learning graph in graph convolutional neural networks for robust seizure prediction,'' \emph{Journal of neural engineering}, vol.~17, no.~3, p. 035004, 2020.

\bibitem{10.3389/fneur.2024.1389731}
\BIBentryALTinterwordspacing
Y.~G. Chung, A.~Cho, H.~Kim, and K.~J. Kim, ``Single-channel seizure detection with clinical confirmation of seizure locations using chb-mit dataset,'' \emph{Frontiers in Neurology}, vol. Volume 15 - 2024, 2024. [Online]. Available: \url{https://www.frontiersin.org/journals/neurology/articles/10.3389/fneur.2024.1389731}
\BIBentrySTDinterwordspacing

\bibitem{vidyaratne2016deep}
L.~Vidyaratne, A.~Glandon, M.~Alam, and K.~M. Iftekharuddin, ``Deep recurrent neural network for seizure detection,'' in \emph{2016 International Joint Conference on Neural Networks (IJCNN)}.\hskip 1em plus 0.5em minus 0.4em\relax IEEE, 2016, pp. 1202--1207.

\bibitem{talathi2017deep}
S.~S. Talathi, ``Deep recurrent neural networks for seizure detection and early seizure detection systems,'' \emph{arXiv preprint arXiv:1706.03283}, 2017.

\bibitem{10.1007/978-981-99-4742-3_25}
J.~Xu, J.~Wang, J.-X. Liu, J.~Shang, L.~Dai, K.~Yan, and S.~Yuan, ``Epileptic seizure detection based on feature extraction and cnn-bigru network with attention mechanism,'' in \emph{Advanced Intelligent Computing Technology and Applications}, D.-S. Huang, P.~Premaratne, B.~Jin, B.~Qu, K.-H. Jo, and A.~Hussain, Eds.\hskip 1em plus 0.5em minus 0.4em\relax Singapore: Springer Nature Singapore, 2023, pp. 308--319.

\bibitem{Lu_2025}
\BIBentryALTinterwordspacing
G.~Lu, J.~Peng, B.~Huang, C.~Gao, T.~Stefanov, Y.~Hao, and Q.~Chen, ``Slimseiz: Efficient channel-adaptive seizure prediction using a mamba-enhanced network,'' in \emph{2025 IEEE International Symposium on Circuits and Systems (ISCAS)}.\hskip 1em plus 0.5em minus 0.4em\relax IEEE, May 2025, p. 1–5. [Online]. Available: \url{http://dx.doi.org/10.1109/ISCAS56072.2025.11043364}
\BIBentrySTDinterwordspacing

\bibitem{9773289}
A.~Chua, M.~I. Jordan, and R.~Muller, ``Soul: An energy-efficient unsupervised online learning seizure detection classifier,'' \emph{IEEE Journal of Solid-State Circuits}, vol.~57, no.~8, pp. 2532--2544, 2022.

\bibitem{Chen_2024}
\BIBentryALTinterwordspacing
Q.~Chen, C.~Sun, C.~Gao, and S.-C. Liu, ``Epilepsy seizure detection and prediction using an approximate spiking convolutional transformer,'' in \emph{2024 IEEE International Symposium on Circuits and Systems (ISCAS)}.\hskip 1em plus 0.5em minus 0.4em\relax IEEE, May 2024, p. 1–5. [Online]. Available: \url{http://dx.doi.org/10.1109/ISCAS58744.2024.10558341}
\BIBentrySTDinterwordspacing

\bibitem{https://doi.org/10.1002/eng2.12918}
\BIBentryALTinterwordspacing
A.~Esmaeilpour, S.~S. Tabarestani, and A.~Niazi, ``Deep learning-based seizure prediction using eeg signals: A comparative analysis of classification methods on the chb-mit dataset,'' \emph{Engineering Reports}, vol.~6, no.~11, p. e12918, 2024. [Online]. Available: \url{https://onlinelibrary.wiley.com/doi/abs/10.1002/eng2.12918}
\BIBentrySTDinterwordspacing

\bibitem{hussein2018epileptic}
R.~Hussein, H.~Palangi, R.~Ward, and Z.~J. Wang, ``Epileptic seizure detection: A deep learning approach,'' \emph{arXiv preprint arXiv:1803.09848}, 2018.

\bibitem{Xiang2025}
\BIBentryALTinterwordspacing
J.~Xiang, Y.~Li, X.~Wu, Y.~Dong, X.~Wen, and Y.~Niu, ``Synchronization-based graph spatio-temporal attention network for seizure prediction,'' \emph{Scientific Reports}, vol.~15, no.~1, p. 4080, 2025. [Online]. Available: \url{https://doi.org/10.1038/s41598-025-88492-5}
\BIBentrySTDinterwordspacing

\bibitem{gui2024eegmambabidirectionalstatespace}
\BIBentryALTinterwordspacing
Y.~Gui, M.~Chen, Y.~Su, G.~Luo, and Y.~Yang, ``Eegmamba: Bidirectional state space model with mixture of experts for eeg multi-task classification,'' 2024. [Online]. Available: \url{https://arxiv.org/abs/2407.20254}
\BIBentrySTDinterwordspacing

\bibitem{mamba}
A.~Gu and T.~Dao, ``Mamba: Linear-time sequence modeling with selective state spaces,'' \emph{arXiv preprint arXiv:2312.00752}, 2023.

\bibitem{gu2021efficiently}
A.~Gu, K.~Goel, and C.~R{\'e}, ``Efficiently modeling long sequences with structured state spaces,'' \emph{arXiv preprint arXiv:2111.00396}, 2021.

\bibitem{auger2021guide}
M.~Auger-M{\'e}th{\'e}, K.~Newman, D.~Cole, F.~Empacher, R.~Gryba, A.~A. King, V.~Leos-Barajas, J.~Mills~Flemming, A.~Nielsen, G.~Petris \emph{et~al.}, ``A guide to state--space modeling of ecological time series,'' \emph{Ecological Monographs}, vol.~91, no.~4, p. e01470, 2021.

\bibitem{lim2020study}
H.-i. Lim, ``A study on dropout techniques to reduce overfitting in deep neural networks,'' in \emph{Advanced Multimedia and Ubiquitous Engineering: MUE-FutureTech 2020}.\hskip 1em plus 0.5em minus 0.4em\relax Springer, 2020, pp. 133--139.

\bibitem{vaswani2017attention}
A.~Vaswani, N.~Shazeer, N.~Parmar, J.~Uszkoreit, L.~Jones, A.~N. Gomez, {\L}.~Kaiser, and I.~Polosukhin, ``Attention is all you need. neurips, 2017,'' 2017.

\bibitem{ravichandranadam}
V.~Ravichandran, ``Adam: A method for stochastic optimization.''

\bibitem{goldberger2000physiobank}
A.~L. Goldberger, L.~A. Amaral, L.~Glass, J.~M. Hausdorff, P.~C. Ivanov, R.~G. Mark, J.~E. Mietus, G.~B. Moody, C.-K. Peng, and H.~E. Stanley, ``Physiobank, physiotoolkit, and physionet: Components of a new research resource for complex physiologic signals,'' \emph{Circulation}, vol. 101, no.~23, pp. e215--e220, 2000, available at \url{https://physionet.org/content/chbmit/1.0.0/}.

\bibitem{esch2019mne}
L.~Esch, C.~Dinh, E.~Larson, D.~Engemann, M.~Jas, S.~Khan, A.~Gramfort, and M.~S. H{\"a}m{\"a}l{\"a}inen, ``Mne: software for acquiring, processing, and visualizing meg/eeg data,'' in \emph{Magnetoencephalography}.\hskip 1em plus 0.5em minus 0.4em\relax Springer, 2019, pp. 1--17.

\bibitem{samarpita2023differentiating}
S.~Samarpita, R.~Satpathy, B.~Mishra, and R.~Mishra, ``Differentiating mental stress levels: Analysing machine learning algorithms comparatively for eeg-based mental stress classification using mne-python,'' \emph{Journal of Advanced Zoology}, vol.~44, pp. 2605--2618, 2023.

\bibitem{wfdb2023}
C.~Xie, L.~McCullum, A.~Johnson, T.~Pollard, B.~Gow, and B.~Moody, ``Waveform database software package (wfdb) for python,'' \url{https://doi.org/10.13026/9njx-6322}, PhysioNet, 2023, version 4.1.0.

\bibitem{harris2020array}
C.~R. Harris, K.~J. Millman, S.~J. Van Der~Walt, R.~Gommers, P.~Virtanen, D.~Cournapeau, E.~Wieser, J.~Taylor, S.~Berg, N.~J. Smith \emph{et~al.}, ``Array programming with numpy,'' \emph{nature}, vol. 585, no. 7825, pp. 357--362, 2020.

\bibitem{huang2025self}
Y.~Huang, Y.~Chen, S.~Xu, D.~Wu, and X.~Wu, ``Self-supervised learning with adaptive frequency-time attention transformer for seizure prediction and classification,'' \emph{Brain Sciences}, vol.~15, no.~4, p. 382, 2025.

\end{thebibliography}
\bibliographystyle{IEEEtran}

\end{document}